\pdfoutput=1
\documentclass{article}
\usepackage[utf8]{inputenc}
\usepackage{color}
\usepackage{epsfig}
\usepackage[margin=1.4in]{geometry}
\usepackage[maxbibnames=99,backend=bibtex]{biblatex}
\bibliography{egbib.bib}
\usepackage{tabularx}
\usepackage{bbold}
\usepackage{graphicx}
\usepackage{caption}
\usepackage{subcaption}
\usepackage[utf8]{inputenc} % allow utf-8 input
\usepackage[T1]{fontenc}    % use 8-bit T1 fonts
\usepackage{hyperref}       % hyperlinks
\usepackage{url}            % simple URL typesetting
\usepackage{booktabs}       % professional-quality tables
\usepackage{amsfonts}       % blackboard math symbols
\usepackage{nicefrac}       % compact symbols for 1/2, etc.
\usepackage{microtype}      % microtypography

\usepackage[utf8]{inputenc}
\usepackage{esvect}
\usepackage{amsmath}
\usepackage{comment}
\usepackage{enumerate}  
\usepackage{paralist}
\usepackage{graphicx}
\usepackage[utf8]{inputenc} % allow utf-8 input
\usepackage[T1]{fontenc}    % use 8-bit T1 fonts
\usepackage{hyperref}       % hyperlinks
\usepackage{url}            % simple URL typesetting
\usepackage{booktabs}       % professional-quality tables
\usepackage{amsfonts}       % blackboard math symbols
\usepackage{nicefrac}       % compact symbols for 1/2, etc.
\usepackage{microtype}      % microtypography
\usepackage{amsmath}
\usepackage{listings}
\usepackage{color}

\newcommand{\myparagraph}[1]{\vspace{6pt}\noindent{\bf #1}}
\usepackage{multirow}

\newcommand{\ie}{\textit{i}.\textit{e}.,~}
\newcommand{\eg}{\textit{e}.\textit{g}.,~}

\definecolor{codegreen}{rgb}{0,0.6,0}
\definecolor{codegray}{rgb}{0.5,0.5,0.5}
\definecolor{codepurple}{rgb}{0.58,0,0.82}
\definecolor{backcolour}{rgb}{0.95,0.95,0.92}
\definecolor{mygreen}{RGB}{28,172,0} % color values Red, Green, Blue
\definecolor{mylilas}{RGB}{170,55,241}

\usepackage{algorithm}
\usepackage{algpseudocode}
\usepackage{pifont}
\usepackage{authblk}

\algblock{ParFor}{EndParFor}
\algnewcommand\algorithmicparfor{\textbf{parfor}}
\algnewcommand\algorithmicpardo{\textbf{do}}
\algnewcommand\algorithmicendparfor{\textbf{end\ parfor}}
\algrenewtext{ParFor}[1]{\algorithmicparfor\ #1\ \algorithmicpardo}
\algrenewtext{EndParFor}{\algorithmicendparfor}

\usepackage{tikz}
\usetikzlibrary{fit,positioning,calc,shapes.misc}

\algnewcommand{\Inputs}[1]{%
  \State \textbf{Inputs:}
  \Statex \hspace*{\algorithmicindent}\parbox[t]{.8\linewidth}{\raggedright #1}
}
\algnewcommand{\Initialize}[1]{%
  \State \textbf{Initialize:}
  \Statex \hspace*{\algorithmicindent}\parbox[t]{.8\linewidth}{\raggedright #1}
}

\begin{document}
\title{\textbf{Predicting Driver Attention in Critical Situations}}
\author[]{Ye Xia}
\author[]{Danqing Zhang}
\author[]{Jinkyu Kim}
\author[]{Ken Nakayama}
\author[]{Karl Zipser}
\author[]{David Whitney}

\affil[]{University of California, Berkeley}

\affil[ ]{\textit{\{yexia,danqing0703,jinkyu.kim,karlzipser,dwhitney\}@berkeley.edu\},nakayama@g.harvard.edu}}    

% \author{Ye Xia\inst{} \and
% Danqing Zhang\inst{} \and
% Jinkyu Kim\inst{} \and\\
% Ken Nakayama\inst{} \and
% Karl Zipser\inst{} \and
% David Whitney\inst{}}

% \institute{University of California, Berkeley, CA 94720, USA\\
% Correspondence: \email{yexia@berkeley.edu}}

\setcounter{Maxaffil}{0}
\renewcommand\Affilfont{\itshape\small}
\maketitle

%===========================================================
\begin{abstract}
Robust driver attention prediction for critical situations is a challenging computer vision problem, yet essential for autonomous driving. Because critical driving moments are so rare, collecting enough data for these situations is difficult with the conventional in-car data collection protocol---tracking eye movements during driving. Here, we first propose a new in-lab driver attention collection protocol and introduce a new driver attention dataset, Berkeley DeepDrive Attention (BDD-A) dataset, which is built upon braking event videos selected from a large-scale, crowd-sourced driving video dataset. We further propose Human Weighted Sampling (HWS) method, which uses human gaze behavior to identify crucial frames of a driving dataset and weights them heavily during model training. With our dataset and HWS, we built a driver attention prediction model that outperforms the state-of-the-art and demonstrates sophisticated behaviors, like attending to crossing pedestrians but not giving false alarms to pedestrians safely walking on the sidewalk. Its prediction results are nearly indistinguishable from ground-truth to humans. Although only being trained with our in-lab attention data, the model also predicts in-car driver attention data of routine driving with state-of-the-art accuracy. This result not only demonstrates the performance of our model but also proves the validity and usefulness of our dataset and data collection protocol.

\end{abstract}
%===========================================================
\section{Introduction}
Human visual attention enables drivers to quickly identify and locate potential risks or important visual cues across the visual field, such as a darting-out pedestrian, an incursion of a nearby cyclist or a changing traffic light. Drivers' gaze behavior has been studied as a proxy for their attention. Recently, a large driver attention dataset of routine driving \cite{dreyeve2016} has been introduced and neural networks \cite{palazzi2017learning,tawari2017computational} have been trained end-to-end to estimate driver attention, mostly in lane-following and car-following situations. Nonetheless, datasets and prediction models for driver attention in rare and critical situations are still needed.

However, it is nearly impossible to collect enough driver attention data for crucial events with the conventional in-car data collection protocol, \ie collecting eye movements from drivers during driving. This is because the vast majority of routine driving situations consist of simple lane-following and car-following. In addition, collecting driver attention in-car has two other major drawbacks. (i) Single focus: at each moment the eye-tracker can only record one location that the driver is looking at, while the driver may be attending to multiple important objects in the scene with their covert attention, \ie the ability to fixate one's eyes on one object while attending to another object \cite{cavanagh2005tracking}. (ii) False positive gazes: human drivers also show eye movements to driving-irrelevant regions, such as sky, trees, and buildings \cite{palazzi2017learning}. It is challenging to separate these false positives from gazes that are dedicated to driving.

An alternative that could potentially address these concerns is showing selected driving videos to drivers in the lab and collecting their eye movements with repeated measurements while they perform a proper simulated driving task. Although this third-person driver attention collected in the lab is inevitably different from the first-person driver attention in the car, it can still potentially reveal the regions a driver should look at in that particular driving situation from a third-person perspective. These data are greatly valuable for identifying risks and driving-relevant visual cues from driving scenes. To date, a proper data collection protocol of this kind is still missing and needs to be formally introduced and tested. 

Another challenge for driver attention prediction, as well as for other driving-related machine learning problems, is that the actual cost of making a particular prediction error is unknown. Attentional lapses while driving on an empty road does not cost the same as attentional lapses when a pedestrian darts out. Since current machine learning algorithms commonly rely on minimizing average prediction error, the critical moments, where the cost of making an error is high, need to be properly identified and weighted.

%----------------------- Placed for better location
\begin{figure}[t]
\centering
\includegraphics[width=\linewidth]{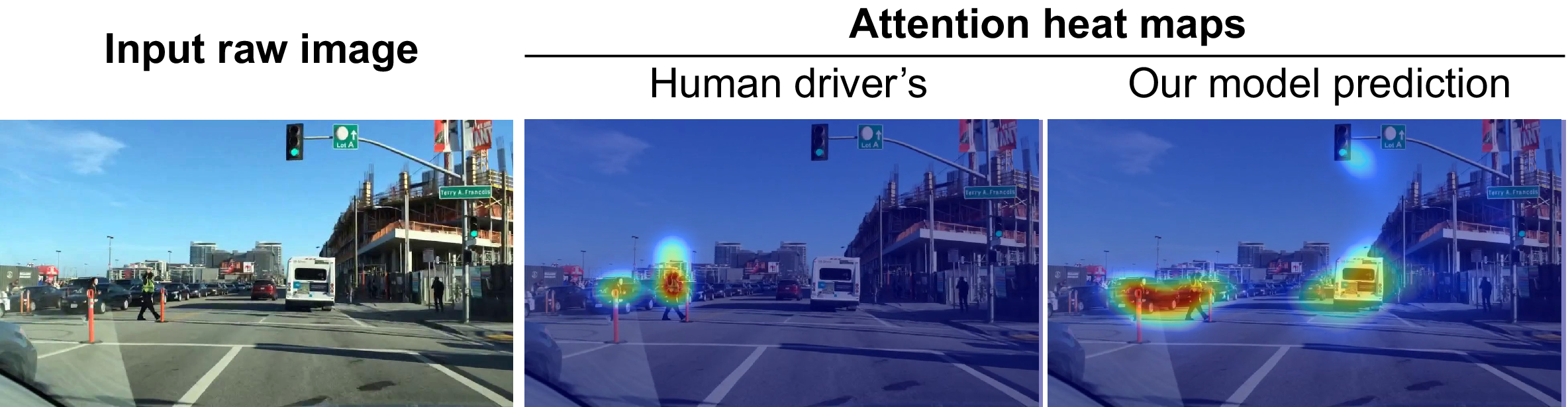}
\caption{An example of input raw images ({\it left}), ground-truth human attention maps collected by us ({\it middle}), and the attention maps predicted by our model ({\it right}). The driver had to sharply stop at the green light to avoid hitting two pedestrians running the red light. The collected human attention map accurately shows the multiple regions that simultaneously demand the driver's attention. Our model correctly attends to the crossing pedestrians and does not give false alarms to other irrelevant pedestrians}
\label{fig:intro}
\end{figure}

Here, our paper offers the following novel contributions. First, in order to overcome the drawbacks of the conventional in-car driver attention collection protocol, we introduce a new protocol that uses crowd-sourced driving videos containing interesting events and makes multi-focus driver attention maps by averaging gazes collected from multiple human observers in lab with great accuracy (Fig.~\ref{fig:intro}). We will refer to this protocol as the in-lab driver attention collection protocol. We show that data collected with our protocol reliably reveal where a experienced driver should look and can serve as a substitute for data collected with the in-car protocol. We use our protocol to collect a large driver attention dataset of braking events, which is, to the best of our knowledge, the richest to-date in terms of the number of interactions with other road agents. We call this dataset Berkeley DeepDrive Attention (BDD-A) dataset and will make it publicly available. Second, we introduce Human Weighted Sampling (HWS), which uses human driver eye movements to identify which frames in the dataset are more crucial driving moments and weights the frames according to their importance levels during model training. We show that HWS improve model performance on both the entire testing set and the subset of crucial frames. Third, we propose a new driver attention prediction model trained on our dataset with HWS. The model shows sophisticated behaviors such as picking out pedestrians suddenly crossing the road without being distracted by the pedestrians safely walking in the same direction as the car (Fig.~\ref{fig:intro}). The model prediction is nearly indistinguishable from ground-truth based on human judges, and it also matches the state-of-the-art performance level when tested on an existing in-car driver attention dataset collected during driving.

\section{Related works}

\myparagraph{Image / Video Saliency Prediction: }
A large variety of the previous saliency studies explored different bottom-up feature-based models \cite{bruce2006saliency,valenti2009image,erdem2013visual,murray2011saliency,zhang2013saliency,bruce2009saliency} combining low-level features like contrast, rarity, symmetry, color, intensity and orientation, or topological structure from a scene \cite{zhang2013saliency,harel2007graph,wei2012geodesic}. Recent advances in deep learning have achieved a considerable improvement for both image saliency prediction \cite{kummerer2014deep,huang2015salicon,liu2015predicting,kummerer2016deepgaze} and video saliency prediction \cite{bazzani2016recurrent,cornia2016predicting,liu2017predicting}. These models have achieved start-of-the-art performance on visual saliency benchmarks collected mainly when human subjects were doing a free-viewing task, but models that are specifically trained for predicting the attention of drivers are still needed.

\myparagraph{Driver Attention Datasets: }
DR(eye)VE \cite{dreyeve2016} is the largest and richest existing driver attention dataset. It contains 6 hours of driving data, but the data was collected from only 74 rides, which limits the diversity of the dataset. In addition, the dataset was collected in-car and has the drawbacks we introduced earlier, including missing covert attention, false positive gaze, and limited diversity. The driver's eye movements were aggregated over a small temporal window to generate an attention map for a frame, so that multiple important regions of one scene might be annotated. But there was a trade-off between aggregation window length and gaze location accuracy, since the same object may appear in different locations in different frames. Reference \cite{fridman2016driver} is another large driver attention dataset, but only six coarse gaze regions were annotated and the exterior scene was not recorded. References \cite{simon2009alerting} and \cite{underwood2011decisions} contain accurate driver attention maps made by averaging eye movements collected from human observers in-lab with simulated driving tasks. But the stimuli were static driving scene images and the sizes of their datasets are small (40 frames and 120 frames, respectively).

\myparagraph{Driver Attention Prediction: }
Self-driving vehicle control has made notable progress in the last several years. One of major approaches is a mediated perception-based approach -- a controller depends on recognizing human-designated features, such as lane markings, pedestrians, or vehicles. Human driver’s attention provides important visual cues for driving, and thus efforts to mimic human driver’s attention have increasingly been introduced. Recently, several deep neural models have been utilized to predict where human drivers should pay attention~\cite{palazzi2017learning,tawari2017computational}. Most of existing models were trained and tested on the DR(eye)VE dataset~\cite{dreyeve2016}. While this dataset is an important contribution, it contains sparse driving activities and limited interactions with other road users. Thus it is restricted in its ability to capture diverse human attention behaviors. Models trained with this dataset tend to become vanishing point detectors, which is undesirable for modeling human attention in urban driving environment, where drivers encounter traffic lights, pedestrians, and a variety of other potential cues and obstacles. In this paper, we provide our human attention dataset as a contribution collected from a publicly available large-scale crowd-sourced driving video dataset~\cite{xu2016end}, which contains diverse driving activities and environments, including lane following, turning, switching lanes, and braking in cluttered scenes.

\section{Berkeley DeepDrive Attention (BDD-A) Dataset}
%\subsection{Data collection and dataset statistics}
\myparagraph{Dataset Statistics: } The statistics of our dataset are summarized and compared with the largest existing dataset (DR(eye)VE) \cite{dreyeve2016} in Table~\ref{tab:dataset_statistics}. Our dataset was collected using videos selected from a publicly available, large-scale, crowd-sourced driving video dataset, BDD100k~\cite{xu2016end,yu2018bdd100k}. BDD100K contains human-demonstrated dashboard videos and time-stamped sensor measurements collected during urban driving in various weather and lighting conditions. To efficiently collect attention data for critical driving situations, we specifically selected video clips that both included braking events and took place in busy areas (see supplementary materials for technical details). We then trimmed videos to include 6.5 seconds prior to and 3.5 seconds after each braking event. It turned out that other driving actions, \eg turning, lane switching and accelerating, were also included. 1,232 videos (=3.5 hours) in total were collected following these procedures. Some example images from our dataset are shown in Fig. \ref{fig:examples}. Our selected videos contain a large number of different road users. We detected the objects in our videos using YOLO~\cite{redmon2017yolo9000}.%, a state-of-the-art object detection algorithm. 
On average, each video frame contained 4.4 cars and 0.3 pedestrians, multiple times more than the DR(eye)VE dataset (Table~\ref{tab:dataset_statistics}).

\begin{table}
    \caption{Comparison between driver attention datasets}
    \label{tab:dataset_statistics}
	\begin{center}
    	\resizebox{\linewidth}{!}{%
    	\begin{tabular}{@{}lccccccc@{}} \toprule
        	\multirow{2}{*}{\parbox{1.5cm}{Dataset}} & \multirow{2}{*}{\parbox{1.5cm}{\centering \# Rides}} & \multirow{1}{*}{\parbox{1.5cm}{\centering Durations}}  & \multirow{2}{*}{\parbox{1.5cm}{\centering \# Drivers}} & \multirow{2}{*}{\parbox{1.5cm}{\centering \# Gaze providers}}    & \multirow{1}{*}{\parbox{1.5cm}{\centering \# Cars}} & \multirow{1}{*}{\# Pedestrians} & \multirow{2}{*}{\parbox{2cm}{\centering \# Braking events}} \\ 
            & & (hours) & & & (per frame) & (per frame) & \\ \midrule
DR(eye)VE~\cite{dreyeve2016} & 74    & 6   & 8       & 8              & 1.0 & 0.04  & 464            \\
BDD-A    & 1,232 & 3.5 & 1,232   & 45             & 4.4 & 0.25  & 1,427          \\\bottomrule
        \end{tabular}}
        
    \end{center}
\end{table} 

%\begin{table}[]
%\centering
%\caption{Comparison between driver attention datasets}
%\label{tab:dataset_statistics}
%\begin{tabular}{cccccccc}
%\hline
%Dataset & Rides & Drivers & Gaze providers & Duration  & Cars      & Pedestrians & Braking events \\ \hline
%DR(eye)VE & 74    & 8       & 8              & 6 hours   & 1.0/frame & 0.04/frame  & 464            \\
%BDDG    & 1,232 & 1,232   & 45             & 3.5 hours & 4.4/frame & 0.25/frame  & 1,427          \\ \hline
%\end{tabular}
%\end{table}

\myparagraph{Data Collection Procedure: } For our eye-tracking experiment, we recruited 45 participants who each had more than one year of driving experience. The participants watched the selected driving videos in the lab while performing a driving instructor task: participants were asked to imagine that they were driving instructors sitting in the copilot seat and needed to press the space key whenever they felt it necessary to correct or warn the student driver of potential dangers. Their eye movements during the task were recorded at 1000 Hz with an EyeLink 1000 desktop-mounted infrared eye tracker, used in conjunction with the Eyelink Toolbox scripts~\cite{cornelissen2002eyelink} for MATLAB. Each participant completed the task for 200 driving videos. Each driving video was viewed by at least 4 participants. The gaze patterns made by these independent participants were aggregated and smoothed to make an attention map for each frame of the stimulus video (see Fig.~\ref{fig:examples} and supplementary materials for technical details).

\begin{figure}
\begin{center}
\includegraphics[width=\linewidth]{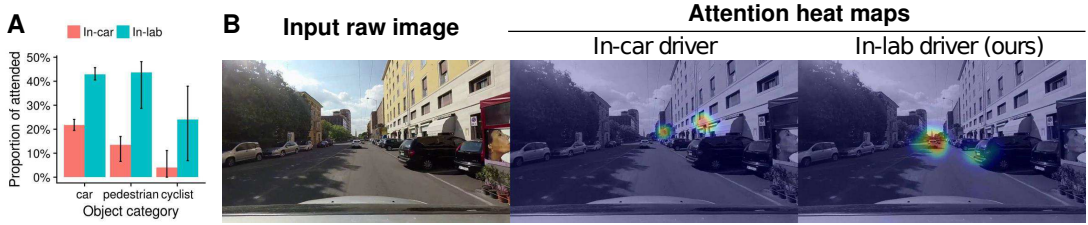}
\end{center}
\caption{Comparison between in-car and in-lab driver attention maps. (A) Proportions of attended objects of different categories for in-car and in-lab driver attention maps. In-car attention maps tend to highlight significantly fewer driving-relevant objects than in-lab attention maps. (B) An example of in-car driver attention maps showing irrelevant regions. The in-lab attention map highlights the car in front and a car that suddenly backed up, while the in-car attention map highlights some regions of the building}
\label{fig:incar_inlab}
\end{figure}

Psychological studies \cite{mannan1997fixation,groner1984looking} have shown that when humans look through multiple visual cues that simultaneously demand attention, the order in which humans look at those cues is highly subjective. Therefore, by aggregating gazes of independent observers, we could record multiple important visual cues in one frame. In addition, it has been shown that human drivers look at buildings, trees, flowerbeds, and other unimportant objects non-negligibly frequently \cite{dreyeve2016}. Presumably, these eye movements should be regarded as noise for driving-related machine learning purposes. By averaging the eye movements of independent observers, we were able to effectively wash out those sources of noise (see Fig.~\ref{fig:incar_inlab}B).

\myparagraph{Comparison with In-Car Attention Data: } We collected in-lab driver attention data using videos from the DR(eye)VE dataset. This allowed us to compare in-lab and in-car attention maps of each video. The DR(eye)VE videos we used were 200 randomly selected 10-second video clips, half of them containing braking events and half without braking events. 

We tested how well in-car and in-lab attention maps highlighted driving-relevant objects. We used YOLO~\cite{redmon2017yolo9000} to detect the objects in the videos of our dataset. We identified three object categories that are important for driving and that had sufficient instances in the videos (car, pedestrian and cyclist). We calculated the proportion of attended objects out of total detected instances for each category for both in-lab and in-car attention maps (see supplementary materials for technical details). The results showed that in-car attention maps highlighted significantly less driving-relevant objects than in-lab attention maps (see Fig.~\ref{fig:incar_inlab}A).

The difference in the number of attended objects between the in-car and in-lab attention maps can be due to the fact that eye movements collected from a single driver do not completely indicate all the objects that demand attention in the particular driving situation. One individual's eye movements are only an approximation of their attention \cite{rizzolatti1987reorienting}, and humans can also track objects with covert attention without looking at them \cite{cavanagh2005tracking}. The difference in the number of attended objects may also reflect the difference between first-person driver attention and third-person driver attention. It may be that the human observers in our in-lab eye-tracking experiment also looked at objects that were not relevant for driving. We ran a human evaluation experiment to address this concern.

\myparagraph{Human Evaluation: } To verify that our in-lab driver attention maps highlight regions that should indeed demand drivers' attention, we conducted an online study to let humans compare in-lab and in-car driver attention maps. In each trial of the online study, participants watched one driving video clip three times: the first time with no edit, and then two more times in random order with overlaid in-lab and in-car attention maps, respectively. The participant was then asked to choose which heatmap-coded video was more similar to where a good driver would look. In total, we collected 736 trials from 32 online participants. We found that our in-lab attention maps were more often preferred by the participants than the in-car attention maps (71\% versus 29\% of all trials, statistically significant as $p$ = 1$\times$10$^{-29}$, see Table~\ref{tab:human_evaluation}). Although this result cannot suggest that in-lab driver attention maps are superior to in-car attention maps in general, it does show that the driver attention maps collected with our protocol represent where a good driver should look from a third-person perspective.

% \begin{table}[b]
% \centering
% \caption{Two human evaluation studies conducted to compare in-lab human driver attention maps with in-car human driver attention maps and attention maps predicted by our HWS model, respectively. In-car human driver attention maps were preferred in significantly less trials than the in-lab human driver attention maps. The attention maps predicted by our HWS model were not preferred in as many trials as the in-lab human driver attention maps, but they achieved significantly higher preference rate than the in-car human driver attention maps.}
% \label{tab:human_evaluation}
% \begin{tabular}{lcc}
% \hline
%                                   & \multicolumn{2}{l}{Preferred by humans over in-lab human driver attention map} \\
%                                   & \# trials                        & proportion of trials                        \\ \hline
% in-car human driver attention map & 216                              & 29\%                                        \\
% HWS model predicted attention map & 189                              & 41\%                                        \\ \hline
% \end{tabular}
% \end{table}

\begin{table}
    \caption{Two human evaluation studies were conducted to compare in-lab human driver attention maps with in-car human driver attention maps and attention maps predicted by our HWS model, respectively. In-car human driver attention maps were preferred in significantly less trials than the in-lab human driver attention maps. The attention maps predicted by our HWS model were not preferred in as many trials as the in-lab human driver attention maps, but they achieved significantly higher preference rate than the in-car human driver attention maps}
    \label{tab:human_evaluation}
	\begin{center}
    	\resizebox{0.5\linewidth}{!}{%
    	\begin{tabular}{@{}cclc@{}} \toprule
            & \# trials & Attention maps & Preference rate \\ \midrule 
            \multirow{2}{*}{Study 1}& \multirow{2}{*}{736}& in-car human driver& 29\%\\ \cmidrule{3-4}
            & & in-lab human driver& 71\%\\ \midrule
            \multirow{2}{*}{Study 2}& \multirow{2}{*}{462}& HWS model predicted& 41\%\\ \cmidrule{3-4}
            & & in-lab human driver& 59\%\\ \bottomrule
        \end{tabular}}
        
    \end{center}
\end{table}

In addition, we will show in the Experiments section that in-lab attention data collected using our protocol can be used to train a model to effectively predict actual, in-car driver attention. This result proves that our dataset can also serve as a substitute for in-car driver attention data, especially in crucial situations where in-car data collection is not practical.

To summarize, compared with driver attention data collected in-car, our dataset has three clear advantages: multi-focus, little driving-irrelevant noise, and efficiently tailored to crucial driving situations.

\section{Attention Prediction Model}
\subsection{Network Configuration}

\begin{figure}[t!]
\centering
\includegraphics[width=\linewidth]{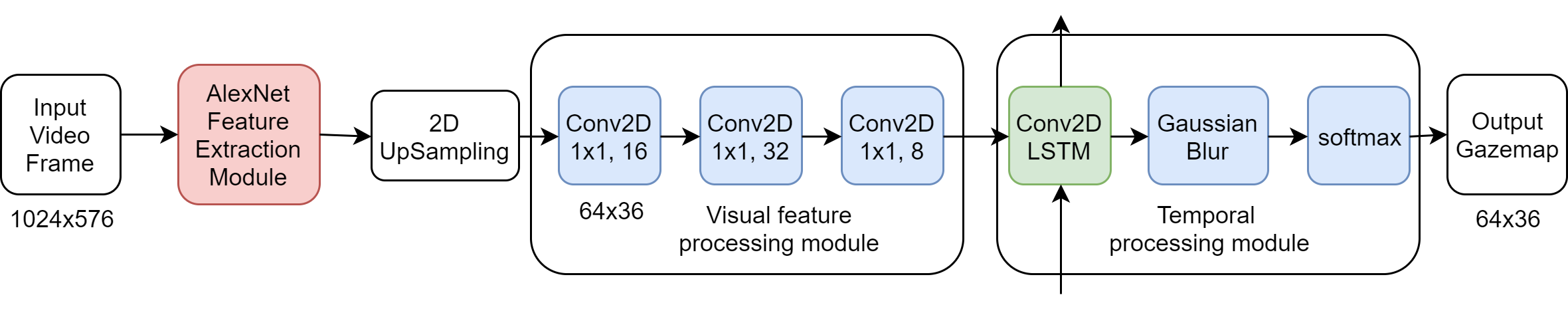} %
\caption{An overview of our proposed model that predicts human driver’s attention from input video frame. We use AlexNet pre-trained on ImageNet as a visual feature extractor. We also use three fully convolutional layers (Conv2D) followed by a convolutional LSTM network (Conv2D LSTM)}
 \label{fig:structure}
\centering
\end{figure}

Our goal is to predict the driver attention map for a video frame given the current and previous video frames. Our model structure can be divided into a visual feature extraction module, a visual feature processing module, and a temporal processing module (Fig.~\ref{fig:structure}).   

The visual feature extraction module is a pre-trained dilated fully convolutional neural network, and its weights are fixed during training. We used ImageNet pre-trained AlexNet \cite{krizhevsky2012imagenet} as our visual feature extraction module. We chose to use the features from the conv5 layer. In our experiment, the size of the input was set to $1024\times576$ pixels, and the feature map by AlexNet was upsampled to $64\times36$ pixels and then fed to the following visual feature processing module.

The visual feature processing module is a fully convolutional neural network. It consists of three convolutional layers with $1 \times 1$ kernels and a dropout layer after each convolutional layer. It further processes the visual features from the previous extraction module and reduces the dimensionality of the visual features from 256 to 8. In our experiments, we observed that without the dropout layers, the model easily got stuck in a suboptimal solution which simply predicted a central bias map, i.e. an attention map concentrated in a small area around the center of the frame.

The temporal processor is a convolutional LSTM network with a kernel size of $3\times3$ followed by a Gaussian smooth layer ($\sigma$ set to 1.5) and a softmax layer. It receives the visual features of successive video frames in sequence from the visual feature processing module and predicts an attention map for every new time step. Dropout is used for both the linear transformation of the inputs and the linear transformation of the recurrent states. We had also experimented with using an LSTM network for this module and observed that the model tended to incorrectly attend to only the central region of the video frames. The final output of this model is a probability distribution over $64\times36$ grids predicting how likely each region of the video frame is to be looked at by human drivers. Cross-entropy is chosen as the loss function to match the predicted probability distribution to the ground-truth.

\begin{figure}[t!]
\centering
\includegraphics[height=3cm]{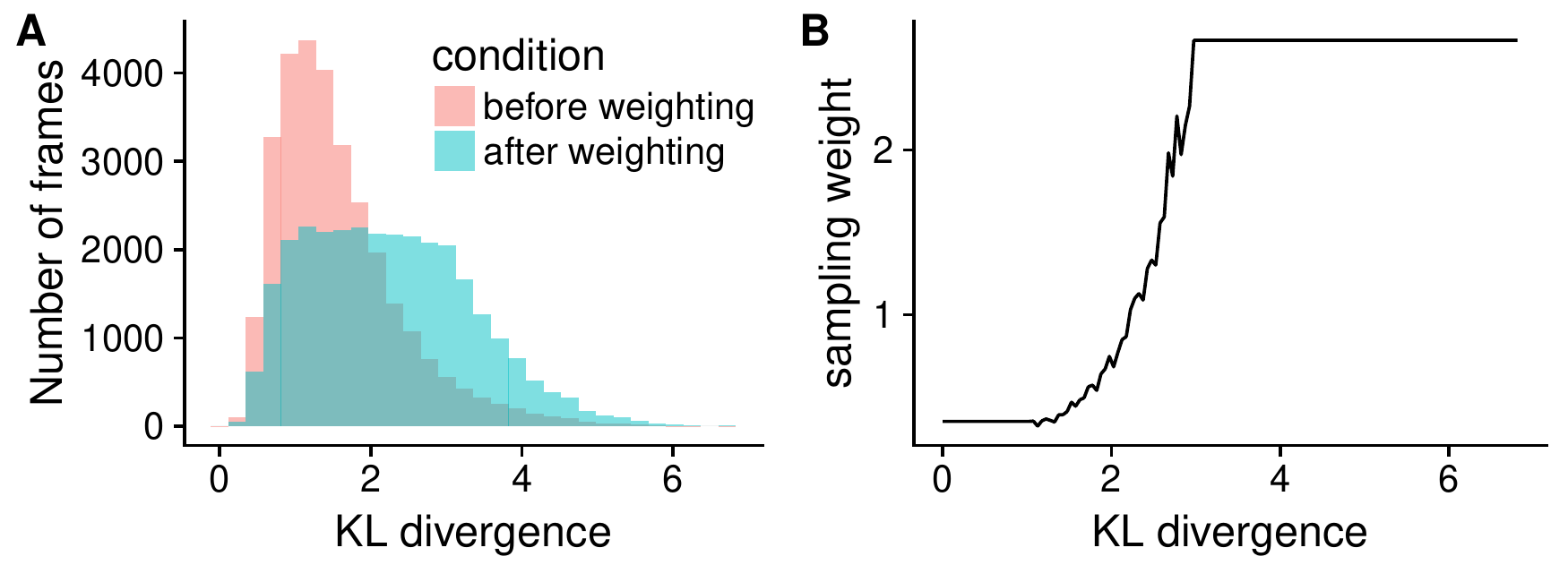} %
\caption{Human Weighted Sampling: (A) For each video frame, we measure the KL divergence between the collected driver attention maps and the mean attention map for that entire video clip ($\approx$10s). We use this computed KL divergence as a weight value to sample image frames during training phase, \ie training a model more often with uncommon attention maps. Histograms show that more uncommon attention maps were selected for training the model, \eg seeing pedestrians or traffic lights is weighted more than just seeing the vanishing point of roads. (B) Normalized sampling weights as a function of KL divergence values. A normalized sampling weight value of 1 indicates that the video frame is sampled once on average during a single epoch}
 \label{fig:kl}
\centering
\end{figure}
%Histogram of KL divergence between individual human attention maps and video average attention maps and sampling weight as a function of the KL divergence value. Panel A: histograms of KL divergence between individual human attention maps of training video frames and their corresponding video average attention maps before and after weighted sampling. The gray vertical lines are showing the KL divergence values of the five example frames illustrated in Figure~\ref{fig:examples}. Panel B: Normalized sampling weight as a function of the KL divergence value. A value of 1 for the normalized sampling weight means the video frame will be on average sampled 1 time when the training algorithm goes through as many training frames as the whole training set.

\subsection{Human Weighted Sampling (HWS)}
Human driver attention datasets, as well as many other driving related datasets, share a common bias: the vast majority of the datasets consist of simple driving situations such as lane-following or car-following. The remaining small proportion of driving situations, such as pedestrians darting out, traffic lights changing, etc., are usually more crucial, in the sense that making errors in these moments would lead to greater cost. Therefore, ignoring this bias and simply using mean prediction error to train and test models can be misleading. In order to tackle this problem, we developed a new method that uses human gaze data to determine the importance of different frames of a driving dataset and samples the frames with higher importance more frequently during training. 

In simple driving situations human drivers only need to look at the center of the road or the car in front, which can be shown by averaging the attention maps of all the frames of one driving video. When the attention map of one frame deviates greatly from the average default attention map, it is usually an important driving situation where the driver has to make eye movements to important visual cues. Therefore, the more an attention map varies from the average attention map of the video, the more important the corresponding training frame is. We used the KL divergence to measure the difference between the attention map of a particular frame and the average attention map of the video. The KL divergence determined the sampling weight of this video frame during training.

The histogram of the KL divergence of all the training video frames of our dataset is shown in Fig.~\ref{fig:kl}. As we expected, the histogram was strongly skewed to the left side. Our goal was to boost up the proportion of the frames of high KL divergence values by weighted sampling. The sampling weight was determined as a function of KL divergence (D$_{\textnormal{KL}}$) illustrated in Fig.~\ref{fig:kl}B. The middle part of this function (D$_{\textnormal{KL}}$ $\in$ [1,3]) was set to be proportional to the inverse of the histogram so that after weighted sampling the histogram of KL divergence would become flat on this range. The left part of the function (D$_{\textnormal{KL}}<$1) was set to a low constant value so that those frames would be sampled occasionally but not completely excluded. The right part of the function was set to a saturated constant value instead of monotonically increasing values in order to avoid over-fitting the model to this small proportion of data. Besides, the attention maps collected in the beginning and the end of each video clip can deviates from the average default attention map merely because the participants were distracted by the breaks between video clips. We therefore restricted the sampling weights of the first second and the last 0.5 seconds of each video to be less or equal to once per epoch. The histogram of KL divergence after weighted sampling is shown in Fig.~\ref{fig:kl}A. In our experiment, we needed to sample the training frames in continuous sequences of 6 frames. For a particular sequence, its sampling weight was equal to the sum of the sampling weights of its member frames. These sequences were sampled at probabilities proportional to the sequence sampling weights.

\section{Results and Discussion}
%We built two driver attention prediction models with the same structure as presented above. One was trained with a regular regime, i.e., equal sampling during training, and the other was trained with HWS. We are going to refer to these two models as our default model and our HWS model. In experiment 1, we compared the performance of our models with a baseline model and a state-of-the-art model in terms of mean prediction errors using some common metrics. In experiment 2, we tested how well the different attention prediction models can pay attention to objects that may be important for driving, e.g., calculating the proportions of attended objects of different categories. In experiment 3, to test how natural and reasonable our model prediction look to humans, we conducted a Turing test to have human observers choose between our model predictions and ground-truth. In experiment 4, to test whether our model trained on in-lab driver attention data can also predict driver attention maps collected in-car, we fine-tuned our HWS model with in-lab attention maps collected using some videos from the DR(eye)VE dataset and tested our model's performance on the in-car attention maps of the DR(eye)VE dataset. In experiment 5, we tested the influence of human driver attention to autonoumous driving models.

Here, we first provide our training and evaluation details, then we summarize the quantitative and qualitative performance comparison with existing gaze prediction models and variants of our model. To test how natural and reasonable our model prediction look to humans, we conduct a human evaluation study and summarize the results. We further test whether our model trained on in-lab driver attention data can also predict driver attention maps collected in-car.

\subsection{Training and Evaluation Details}
We made two variants of our model. One was trained with a regular regime, \ie equal sampling during training, and the other was trained with Human Weighted Sampling (HWS). Except for the sampling method during training, our default model and HWS model shared the same following training settings. We used 926 videos from our BDD-A dataset as the training set and 306 videos as the testing set. We downsampled the videos to $1024 \times 576$ pixels and 3Hz. After this preprocessing, we had about 30k frames in our training set and 10k frames in our testing set. We used cross-entropy between predicted attention maps and human attention maps as the training loss, along with Adam optimizer (learning rate $=0.001$, $\beta_{1}=0.9$, $\beta_{2}=0.999$, $\epsilon=1\times10^{-8}$). Each training batch contained 10 sequences and each sequence had 6 frames. The training was done for 10,000 iterations. The two models showed stabilized testing errors by iteration 10,000.

To our knowledge, \cite{palazzi2017learning} and \cite{tawari2017computational} are the two deep neural models that use dash camera videos alone to predict human driver's gaze. They demonstrated similar results and were shown to surpass other deep learning models or traditional models that predict human gaze in non-driving-specific contexts. We chose to replicate \cite{palazzi2017learning} to compare with our work because their prediction code is public. The model designed by \cite{palazzi2017learning} was trained on the DR(eye)VE dataset \cite{dreyeve2016}. We will refer to \cite{palazzi2017learning}'s model as DR(eye)VE model in the following. The training code of \cite{palazzi2017learning} is not available. We implemented code to fine-tune their model on our dataset, but the fine-tuning did not converge to any reasonable solution, potentially due to some training parameter choices that were not reported. We then tested their pre-trained model directly on our testing dataset without any training on our training dataset. Since the goal of the comparison was to test the effectiveness of the combination of model structure, training data and training paradigm as a whole, we think it is reasonable to test how well DR(eye)VE model performs on our dataset without further training. For further comparison, we fine-tuned a publicly available state-of-the-art image gaze prediction model, SALICON \cite{huang2015salicon} on our dataset. We used the open source implementation \cite{christopherleethomas2016}. We also tested our models against a baseline model that always predicts the averaged human attention map of training videos.

Kullback-Leibler divergence (KL divergence, $D_{KL}$), Pearson’s Correlation Coefficient (CC), Normalized Scanpath Saliency (NSS) and Area under ROC Curve (AUC) are four commonly used metrics for attention map prediction \cite{palazzi2017learning,tawari2017computational,bylinskii2018different}. We calculated the mean prediction errors in these four metrics on the testing set to compare the different models. In order to test how well the models perform at important moments where drivers need to watch out, we further calculated the mean prediction errors on the subset of testing frames where the attention maps deviate significantly from the average attention maps of the corresponding videos (defined as KL divergence greater than 2.0). We will refer to these frames as non-trivial frames. Our models output predicted attention maps in the size of $64 \times 36$ pixels, but the DR(eye)VE model and the SALICON outputs in bigger sizes. For a fair comparison, we scaled the DR(eye)VE model and the SALICON model's predicted attention maps into $64 \times 36$ pixels before calculating the prediction errors.

Another important evaluation criterion of driver attention models is how successfully they can attend to the objects that demand human driver's attention, e.g. the cars in front, the pedestrians that may enter the roadway, etc. Therefore, we applied the same attended object analysis described in the Berkeley DeepDrive Attention Dataset section. We used YOLO~\cite{redmon2017yolo9000} to detect the objects in the videos of our dataset. We selected object categories that are important for driving and that have enough instances in both our dataset and the DR(eye)VE dataset for comparison (car, pedestrian and cyclist). We calculated the proportions of all the detected instances of those categories that were actually attended to by humans versus the models. The technical criterion of determining attended objects was the same as described in the Berkeley DeepDrive Attention Dataset section.

\setlength{\tabcolsep}{4pt}
\begin{table}[t]
    \caption{Performance comparison of human attention prediction. Mean and 95\% bootstrapped confidence interval are reported}
    \label{tab:exp1_metrics}
	\begin{center}
    	\resizebox{1.0\linewidth}{!}{%
    	\begin{tabular}{@{}lcccccccc@{}} \toprule
        \multirow{3}{*}{} & \multicolumn{4}{c}{Entire testing set} & \multicolumn{4}{c}{Testing subset where D$_{\textnormal{KL}}$(GT, Mean) $>$ 2} \\
        & \multicolumn{2}{c}{KL divergence} & \multicolumn{2}{c}{Correlation coefficient} & \multicolumn{2}{c}{KL divergence} & \multicolumn{2}{c}{Correlation coefficient}\\ \cmidrule{2-9}
        & Mean        & 95\% CI             & Mean             & 95\% CI & Mean        & 95\% CI             & Mean             & 95\% CI                   \\ \midrule
Baseline                              & 1.50        & (1.45, 1.54)        & 0.46             & (0.44, 0.48) & 1.87        & (1.80, 1.94)        & 0.36             & (0.34, 0.37)             \\
SALICON~\cite{huang2015salicon}                               & 1.41        & (1.39, 1.44)        & 0.53             & (0.51, 0.54) & 1.76        & (1.72, 1.80)        & 0.39             & (0.37, 0.41)\\
DR(eye)VE~\cite{palazzi2017learning}                             & 1.95        & (1.87, 2.04)        & 0.50             & (0.48, 0.52)            & 2.63        & (2.51, 2.77)        & 0.35             & (0.33, 0.37)             \\
Ours (default)                        & 1.24        & (1.21, 1.28)        & 0.58             & (0.56, 0.59)             & 1.71        & (1.65, 1.79)        & 0.41             & (0.40, 0.43)             \\
Ours (HWS)                            & {\bf{1.24}}        & {\bf{(1.21, 1.27)}}        & {\bf{0.59}}             & {\bf{(0.57, 0.60)}}             & {\bf{1.67}}        & {\bf{(1.61, 1.73)}}        & {\bf{0.44}}             & {\bf{(0.42, 0.45)}}             \\ \bottomrule
        \end{tabular}}
    \end{center}
\end{table} 
\setlength{\tabcolsep}{1.4pt}

\subsection{Evaluating Attention Predictor}
%\subsection{Experiment 1: prediction accuracy in terms of mean predictions}
\myparagraph{Quantitative Analysis of Attention Prediction: } The mean prediction errors of different models are summarized in Table~\ref{tab:exp1_metrics} (measured in $D_{KL}$ and CC) and Table S1 (measured in NSS and AUC) in supplementary materials. Both of our models significantly outperformed the DR(eye)VE model, the SALICON model and the baseline model in all metrics on both the entire testing set and the subset of non-trivial frames. Our model trained with HWS was essentially trained on a dataset whose distribution was altered from the distribution of the testing set. However, our HWS model showed better results than our default model even when being tested on the whole testing set. When being tested on the subset of non-trivial frames, our HWS model outperformed our default model even more significantly. These results suggest that HWS has the power to overcoming the dataset bias and better leveraging the knowledge hidden in crucial driving moments. 

\begin{figure}[t]
\centering
\includegraphics[width=\linewidth]{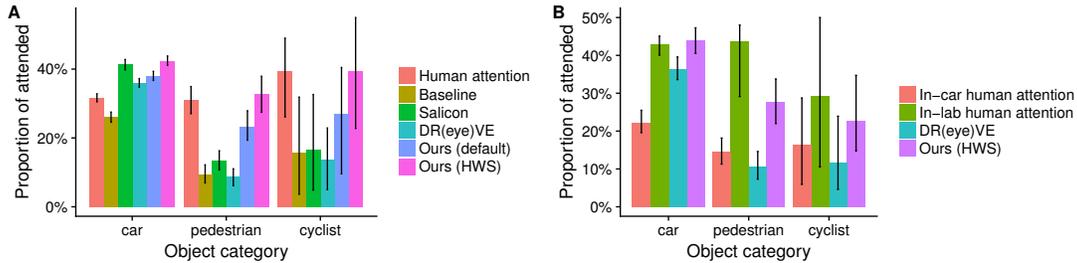}
\caption{Analysis of attended objects for human attention and different models tested on our dataset (A) and the DR(eye)VE dataset (B). Error bars show 95\% bootstrapped confidence intervals}
\label{fig:model_object}
\end{figure}

The results of the attended object analysis are summarized in Fig.~\ref{fig:model_object}A. Cars turned out to be easy to identify for all models. This is consistent with the fact that a central bias of human attention is easy to learn and cars are very likely to appear in the center of the road. However, for pedestrians and cyclists, the DR(eye)VE model, SALICON model and baseline model all missed a large proportion of them compared with human attention ground-truth. Both of our models performed significantly better than all the other competing models in the categories of pedestrians and cyclists, and our HWS model matched the human attention performances the best.

Importantly, our HWS model did not simply select objects according to their categories like an object detection algorithm. Considering the category that has the highest safety priority, pedestrian, our models selectively attended to the pedestrians that were also attended to by humans. Let us refer to the pedestrians that were actually attended to by humans as the important pedestrians and the rest of them as non-important pedestrians. Among all the pedestrians detected by the object detection algorithm, the proportion of important pedestrians was $33\%$. If our HWS model were simply detecting pedestrians at a certain level and could not distinguish between important pedestrians and non-important pedestrians, the proportion of important pedestrians among the pedestrians attended to by our model should also be $33\%$. However, the actual proportion of important pedestrians that our HWS model attended to was  $48\%$ with a bootstrapped $95\%$ confidence interval of $[42\%, 55\%]$. Thus, our HWS model predicts which of the pedestrians are the ones most relevant to human drivers.

%\subsection{Experiment 2: how much proportions can the driver attention models pay attention to important objects?}
\myparagraph{Qualitative Analysis of Attention Prediction: } Some concrete examples are shown in Figure~\ref{fig:examples} (see supplementary information for videos). These examples demonstrates some important driving scenarios: pedestrian crossing, cyclist getting very close to the vehicle and turning at a busy crossing. It can be seen from these examples that the SALICON model and the DR(eye)VE model mostly only predicted to look at the center of the road and ignored the crucial pedestrians or cyclists. In the examples of row 1, 2 and 3, both our default model and HWS model successfully attended to the important pedestrian/cyclist, and did not give false alarm for other pedestrians who were not important for the driving decision. In the challenging example shown in row 4, the driver was making a right turn and needed to yield to the crossing pedestrian. Only our HWS model successfully overcame the central bias and attended to the pedestrian appearing in a quite peripheral area in the video frame.

%The important pedestrians are labeled by red rectangles and the non-important pedestrians are labeled by white. The example in row 1 demonstrated a pedestrian who was about to cross the road while speaking on a phone without looking at the driver. It was crucial to be aware of that pedestrian and stop the car even though the driver had the green light. Our models attended to that pedestrian, but did not attend to the other pedestrian more to the right who was larger in the video frame but not crucial for the driver's decision. In the example in row 2, the driver had a yellow light and some pedestrians were about to enter the roadway. Our models attended to those pedestrians, who informed the driver whether to proceed through the yellow light; our model did not attend to the other (less important) pedestrian. In the example in row 3, our models successfully attended to a cyclist who was very close to the car. The examples in row 4 and 5 are more challenging, and only our HWS model succeeded and demonstrated some sophisticated behaviours. In the example in row 4, our HWS model successfully predicted attending to the person exiting from the parked car, but did not attend to the passenger who exited from the other side of the car. In the example in row 5, the driver was making a right turn and needed to yield to the crossing pedestrian. Even though the pedestrian appeared in a quite peripheral area in that video frame, our HWS model successfully overcame the central bias and attended to that pedestrian.

\begin{figure}[t]
\centering
\includegraphics[width=\linewidth]{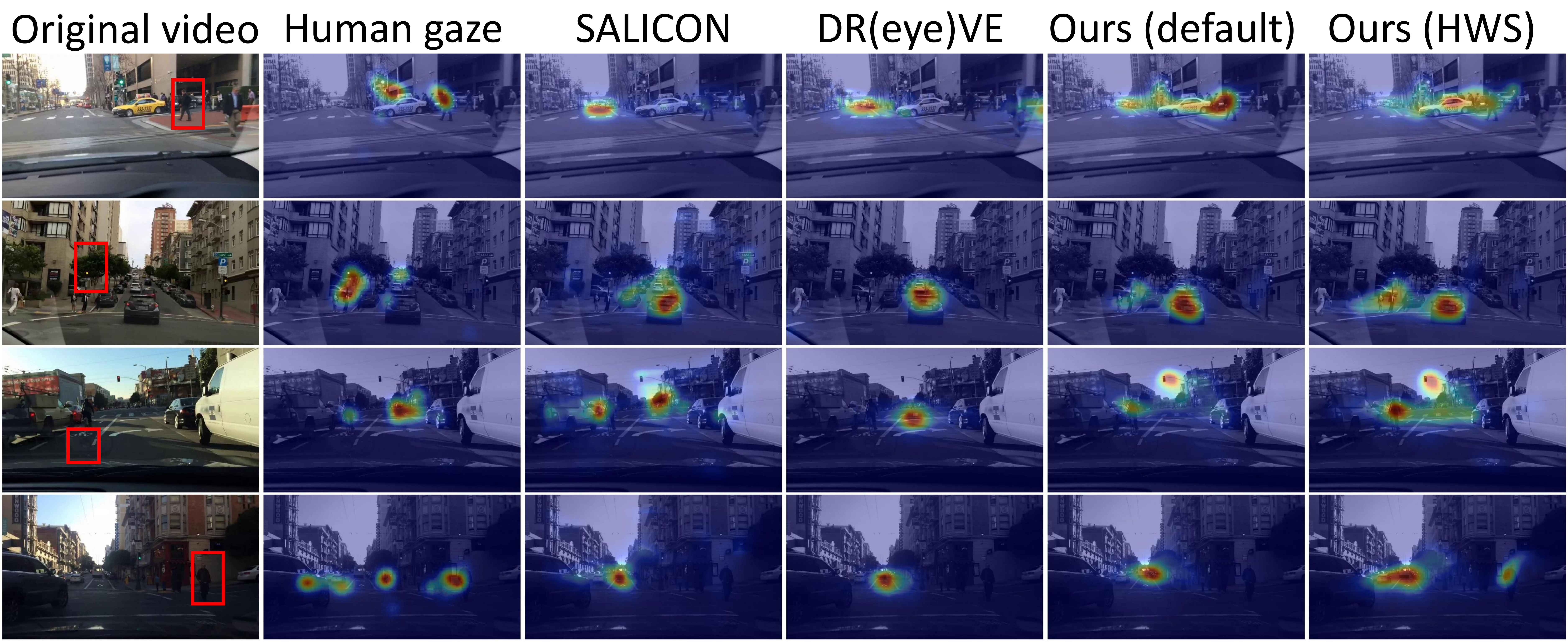}
\caption{Examples of the videos in our dataset, ground-truth human attention maps and the prediction of different models. The red rectangles in the original video column highlight the pedestrians that pose a potential hazard. Row 1: the driver had the green light, but a pedestrian was about to cross the road while speaking on a phone without looking at the driver. Another pedestrian was present in the scene, but not relevant to the driving decision. Row 2: the driver had a yellow light and some pedestrians were about to enter the roadway. Another pedestrian was walking in the same direction as the car and therefore not relevant to the driving decision. Row 3: a cyclist was very close to the car. Row 4: the driver was making a right turn and needed to yield to the crossing pedestrian. Other pedestrians were also present in the scene but not relevant to the driving decision}
\label{fig:examples}
\end{figure}

\myparagraph{Human Evaluation: } To further test how natural and reasonable our HWS model's predicted attention maps look to humans, we conducted an online Turing Test. In each trial, a participant watched one driving video clip three times: the first time with no edit, and then two times in random order with the ground-truth human driver attention map and our HWS model's predicted attention map overlaid on top, respectively. The participant was then asked to choose whether the first or the second attention map video was more similar to where a good driver would look.

Note that the experiment settings and instructions were the same as the online study described in the dataset section, except that one compares model prediction against the in-lab driver attention maps, and the other compares the in-car driver attention maps against the in-lab driver attention maps. Therefore, the result of this Turing Test can be compared with the result of the previous online study. In total, we collected 462 trials from 20 participants. If our HWS model's predicted attention maps were perfect and indistinguishable from the ground-truth human driver attention maps, the participants would had to make random choices, and therefore we would expect them to choose our model prediction in about $50\%$ of the trials. If our HWS model's prediction was always wrong and unreasonable, we would expect a nearly zero chosen rate for our model prediction. Our results showed that in 41\% of all trials the participants chose our HWS model's predicted attention maps as even better than the in-lab human attention maps (see Table~\ref{tab:human_evaluation}). In the previous online study, the in-car attention maps of DR(eye)VE only achieved a chosen rate of 29\%. This result suggests that our HWS model's predicted attention maps were even more similar to where a good driver should look than the human driver attention maps collected in-car (permutation test p = $4 \times 10^{-5}$).

\subsection{Predicting In-Car Driver Attention Data:}
To further demonstrate that our model has good generalizability and that our driver attention data collected in-lab is realistic, we conducted a challenging test: we trained our model using only our in-lab driver attention data, but tested it on the DR(eye)VE dataset, an in-car driver attention dataset. Note that the DR(eye)VE dataset covers freeway driving, which is not included in our dataset due to the small density of road user interactions on freeway. The high driving speed on freeway introduces strong motion blur which is not present in our dataset videos. Furthermore, drivers need to look further ahead in high speed situations, so the main focus of driver gaze pattern shifts up as the driving speed increases. In order to adapt our model to these changes, we selected 200 ten-second-long video clips from the training set of the DR(eye)VE dataset and collected in-lab driver attention maps for those video clips (already described in the Berkeley DeepDrive Attention Dataset section). We fine-tuned our HWS model with these video clips (30 minutes in total only) and the corresponding in-lab driver attention maps, and then tested the model on the testing set of the DR(eye)VE dataset (with in-car attention maps). The mean testing errors were calculated in $D_{KL}$ and CC because the calculation of NSS and AUC requires the original fixation pixels instead of smoothed gaze maps and the original fixation pixels of the DR(EYE)VE dataset were not released. Our fine-tuned model showed a better mean value in KL Divergence and a worse mean value in CC than the DR(eye)VE model (see Table~\ref{tab:exp4_metrics}). But the 95\% bootstrapped confidence intervals for the two models in both metrics overlapped with each other. So overall we concluded that our fine-tuned model matched the performance of the DR(eye)VE model. Note that the DR(eye)VE model was trained using the DR(eye)VE dataset and represents the state-of-the-art performance on this dataset. 

\setlength{\tabcolsep}{4pt}
\begin{table}[t]
    \caption{Test  results  obtained on the DR(eye)VE dataset by  the  state-of-the-art  model (DR(eye)VE) and our finetuned model. Mean and 95\% bootstrapped confidence interval are reported}
    \label{tab:exp4_metrics}
	\begin{center}
    	\resizebox{0.6\linewidth}{!}{%
    	\begin{tabular}{@{}lcccc@{}} \toprule
        \multirow{2}{*}{} & \multicolumn{2}{c}{KL divergence} & \multicolumn{2}{c}{Correlation coefficient} \\ \cmidrule{2-5}
        & Mean        & 95\% CI             & Mean             & 95\%CI                   \\ \midrule
DR(eye)VE                               & 1.76        & (1.65, 1.87)        & 0.54             & (0.51, 0.56)             \\
Ours (finetuned)                      & 1.72        & (1.66, 1.81)        & 0.51             & (0.48, 0.53)             \\ \bottomrule
        \end{tabular}}
        
    \end{center}
\end{table} 
\setlength{\tabcolsep}{1.4pt}

We also calculated proportions of attended objects of important categories for our fine-tuned model and the DR(eye)VE model (Fig.~\ref{fig:model_object}B). Our fine-tuned model showed significantly higher proportions of attended objects in the car, pedestrian and cyclist categories and was more similar to the in-lab driver attention than the DR(eye)VE model. Note that we have shown in the Berkeley DeepDrive Attention Dataset section that humans rated the in-lab attention maps as more similar to where a good driver should look from a third-person perspective than the in-car attention maps.

%------------------------------------------------------------------------- 
\section{Conclusions}
In this paper, we introduce a new in-lab driver attention data collection protocol that overcomes drawbacks of in-car collection protocol. We contribute a human driver attention dataset which is to-date the richest and will be made public. We propose Human Weighted Sampling which can overcome common driving dataset bias and improve model performance in both the entire dataset and the subset of crucial moments. With our dataset and sampling method we contribute a novel human driver attention prediction model that can predict both in-lab and in-car driver attention data. The model demonstrates sophisticated behaviors and show prediction results that are nearly indistinguishable from ground-truth to humans. 

\vspace{5mm}
\myparagraph{Acknowledgement: }
This work was supported by Berkeley DeepDrive. We thank Yuan Yuan and Victor Hiltner for assistance with driver gaze data collection, and Victor Hiltner, Kevin Li and Drew Kaul for comments and proofreading that greatly improved the manuscript.
\vspace{3mm}

%===========================================================
%\bibliographystyle{splncs}
%\bibliography{egbib}
\printbibliography
%this would normally be the end of your paper, but you may also have an appendix
%within the given limit of number of pages
\end{document}